\documentclass[letterpaper]{article} 
\usepackage{aaai2026}  
\usepackage{times}  
\usepackage{helvet}  
\usepackage{courier}  
\usepackage[hyphens]{url}  
\usepackage{graphicx} 
\usepackage{natbib}  
\usepackage{caption} 
\usepackage{algorithm}
\usepackage{algorithmic}
\usepackage{graphicx}%
\usepackage{multirow}%
\usepackage{amsmath,amssymb,amsfonts}%
\usepackage{amsthm}%
\usepackage{mathrsfs}%
\usepackage[title]{appendix}%
\usepackage{xcolor}%
\usepackage{textcomp}%
\usepackage{manyfoot}%
\usepackage{booktabs}%
\usepackage{listings}%
\usepackage{xspace}

\newcommand{\furl}[1]{\footnote{\scriptsize \url{#1}}}
\newcommand{\ftext}[1]{\footnote{\scriptsize #1}}

\newcommand{\ner}[1]{\textsf{#1}\xspace}

\usepackage{newfloat}
\usepackage{listings}
\DeclareCaptionStyle{ruled}{labelfont=normalfont,labelsep=colon,strut=off} 
\floatstyle{ruled}
\newfloat{listing}{tb}{lst}{}
\floatname{listing}{Listing}
\title{Large Language Models for Few-Shot Named Entity Recognition}
\author{
    Yufei Zhao\textsuperscript{\rm 1}, Xiaoshi Zhong\textsuperscript{\rm 1}\thanks{Xiaoshi Zhong is the corresponding author.}, Erik Cambria\textsuperscript{\rm 2}, and Jagath C. Rajapakse\textsuperscript{\rm 2}\\
}
\affiliations{
    \textsuperscript{\rm 1}School of Computer Science and Technology, Beijing Institute of Technology, China.\\
    \textsuperscript{\rm 2}College of Computing and Data Science, Nanyang Technological University, Singapore.

    yfzhao@bit.edu.cn; 
    xszhong@bit.edu.cn; 
    cambria@ntu.edu.sg; 
    asjagath@ntu.edu.sg
    
}

\usepackage{bibentry}

\begin{document}

\maketitle

\begin{abstract}
Named entity recognition (NER) is a fundamental task in numerous downstream applications. Recently, researchers have employed pre-trained language models (PLMs) and large language models (LLMs) to address this task. However, fully leveraging the capabilities of PLMs and LLMs with minimal human effort remains challenging. In this paper, we propose GPT4NER, a method that prompts LLMs to resolve the few-shot NER task. GPT4NER constructs effective prompts using three key components: entity definition, few-shot examples, and chain-of-thought. By prompting LLMs with these effective prompts, GPT4NER transforms few-shot NER, which is traditionally considered as a sequence-labeling problem, into a sequence-generation problem. We conduct experiments on two benchmark datasets, CoNLL2003 and OntoNotes5.0, and compare the performance of GPT4NER to representative state-of-the-art models in both few-shot and fully supervised settings. Experimental results demonstrate that GPT4NER achieves the $F_1$ of 83.15\% on CoNLL2003 and 70.37\% on OntoNotes5.0, significantly outperforming few-shot baselines by an average margin of 7 points. Compared to fully-supervised baselines, GPT4NER achieves 87.9\% of their best performance on CoNLL2003 and 76.4\% of their best performance on OntoNotes5.0. We also utilize a relaxed-match metric for evaluation and report performance in the sub-task of named entity extraction (NEE), and experiments demonstrate their usefulness to help better understand model behaviors in the NER task.
\end{abstract}

\begin{links}
    \link{Code and Data}{https://github.com/xszhong/GPT4NER}
\end{links}

\section{Introduction}

Named entity recognition (NER) \cite{chinchor1997muc} is a fundamental task in natural language processing, aiming to extract and classify named entities from unstructured text. By transforming original text into structured data, NER provides crucial support for many downstream tasks, making its accuracy essential for subsequent tasks. Traditionally, NER is treated as a sequence-labeling task~\citep{devlin-etal-2019-bert, yang-katiyar-2020-simple}. Early methods primarily rely on large annotated corpora from specific domains and employ supervised or semi-supervised learning algorithms to address this task~\citep{yang-katiyar-2020-simple, ding-etal-2020-daga}. While these methods perform well on closed datasets~\citep{wang-etal-2021-automated, Li_Fei_Liu_Wu_Zhang_Teng_Ji_Li_2022}, they often require access to complete labeled training datasets for model training and fail to meet the demands of open-ended business scenarios in industry due to limitations in labeled data and the scarcity of data in specific domains such as biomedicine and materials science. 

In the early stages, the NER task~\citep{chinchor1997muc} primarily relies on rule-based methods~\citep{hanisch2005prominer, 10.5555/1870457.1870476} and dictionary-based methods~\citep{sasaki2008make, 10.1197/jamia.M1453}, which require experts to manually construct rules based on dataset features. This process is both time-consuming and labor-intensive. With the advancement of machine-learning techniques, researchers adopt machine learning-based methods to resolve the NER task. Hidden Markov models (HMMs)~\citep{10.1197/jamia.M1453, morwal2012named, 10.5555/1567594.1567613} and conditional random fields (CRFs)~\citep{xu2008crf, LI2009334} becomes particularly representative of this approach. While these statistical machine learning-based NER models significantly improve the performance, they require extensive manual annotation of domain-specific data, limiting their scalability and practical application. The rise of deep-learning and neural-network techniques further transform the NER task. Researchers employ these methods, with commonly used NER models including convolutional neural networks (CNNs)~\citep{10.5555/1953048.2078186, 10.1162/tacl_a_00104, ma-hovy-2016-end} and recurrent neural networks (RNNs)~\citep{lyu2017long, chowdhury2018multitask}, among others. These deep-learning models can automatically learn feature representations from large-scale data and have achieved significant improvements in the NER task.

\citet{devlin-etal-2019-bert} transfer the pre-trained language model BERT to fine-tuning on 11 natural language processing benchmark tasks, achieving state-of-the-art results. Since then, NER methods have increasingly relied on large-scale pre-trained language models, which leverage benefits of big data and large-scale computing. \citet{wang-etal-2022-deepstruct} propose structural pre-training, which guides language models to generate structures from text and enhances knowledge transfer between different tasks. Context learning has also been applied to the NER task. For example, \citet{chen-etal-2023-learning} design a meta-function pre-training algorithm to inject context learning capabilities into pre-trained language models, which enables rapid identification of new entity types using demonstration instances. Additionally, data augmentation techniques have been used to alleviate the scarcity of labeled data in NER. For example, \citet{hu-etal-2023-entity} propose an entity-to-text data augmentation technique that utilizes pre-trained large-scale language models to construct an augmented entity list.

With the rise and widespread use of large language models (LLMs) such as OpenAI's GPT series (e.g., GPT-3~\citep{10.5555/3495724.3495883} and GPT4~\citep{achiam2023gpt}), various NLP tasks have achieved promising results, including relation extraction~\citep{wadhwa-etal-2023-revisiting, dagdelen2024structured} and question answering~\citep{10.5555/3600270.3600452}. Trained on diverse datasets across multiple domains, LLMs exhibit powerful capabilities in understanding context and generating natural language text. With only a few examples as demonstrations for a specific task, LLMs can generate accurate responses to new inputs. In the era of LLMs, numerous studies have explored their application to NER tasks, including few-shot learning~\citep{huang-etal-2022-copner, wang-etal-2025-gpt, ashok2023promptnerpromptingnamedentity}, zero-shot learning~\citep{xie-etal-2023-empirical, 10.1093/jamia/ocad259, Shao_2024}, fine-tuning models for target domains, and using GPT as a data generator for data augmentation~\citep{10.1145/3539618.3591957, ye2024llmdadataaugmentationlarge}. \citet{zhao2025few} propose a few-shot biomedical NER method that combines LLM-assisted data augmentation with multi-scale feature extraction, effectively improving model performance on multiple biomedical datasets under few-shot settings. Few-shot methods typically include domain transfer and prompt engineering. Domain transfer methods~\citep{das-etal-2022-container,chen-etal-2023-prompt} usually train on large amounts of source data and fine-tuning on examples from target domain. Prompt engineering methods~\citep{wang-etal-2025-gpt,zhou2024universalnertargeteddistillationlarge,10197069} often adopt a strategy of querying for the presence of one specific entity type at a time to improve recognition accuracy. However, this querying approach significantly increases time when dealing with multiple entity types, especially more entity types or longer test text processing. The time cost becomes a critical bottleneck in such cases. 

Chain-of-thought prompting provides statement reasoning and maintains complete interpretability~\citep{10.5555/3600270.3602070}. However, it performs poorly when addressing problems more complex than provided examples. \citet{zhou2023leasttomostpromptingenablescomplex} introduce ``least-to-most prompting'' to decompose a complex problem into a series of sub-problems and address them sequentially, which enables the model to solve problems harder than the examples. \citet{zhang2022automaticchainthoughtprompting} propose the auto-CoT paradigm to automatically constructs questions and reasoning chains, which improves fault tolerance. 

\citet{ashok2023promptnerpromptingnamedentity} apply chain-of-thought prompting to few-shot NER tasks, achieving cross-domain applications and improving flexibility by modifying definitions and examples. However, the few-shot examples are selected randomly without a targeted selection strategy, and evaluation is conducted on a random sample of 500 test examples by reporting mean and variance over 5 runs, which may not reflect performance across the full dataset or multi-type entity scenarios. \citet{wang-etal-2025-gpt} apply GPT-3 to the NER task by converting sequence labeling into a generation task, requiring the identification of entity types after providing prompts and examples (obtaining the nearest neighbors as examples through k-nearest neighbors). They use a self-verification strategy to address the hallucination problem of LLMs. However, this method can only extract one type of entities at a time. In datasets with many entity types, this can result in more time spent. \citet{zhou2024universalnertargeteddistillationlarge} compare querying all types of entities at once to querying one type of entity at a time, the mode of querying all types of entities at once is not efficient. \citet{guo-etal-2025-baner} propose BANER, a boundary-aware NER framework leveraging contrastive learning and LoRAHub for cross-domain adaptation. While BANER improves entity boundary detection in few-shot settings, it employs a single, stage-specific prompt template for each phase, which may limit flexibility and expressiveness. 

Inspired by these LLM-based methods such as PromptNER \cite{ashok2023promptnerpromptingnamedentity}, GPT-NER \cite{wang-etal-2025-gpt}, and BANER \cite{guo-etal-2025-baner}, in this paper, we propose GPT4NER, an LLM-based method that leverages the capabilities of LLMs to tackle the few-shot NER task. GPT4NER enables querying for all entity types in a single query, reducing querying time. GPT4NER constructs effective prompts using three key components: entity definition, few-shot examples, and chain-of-thought, with an optional component of part-of-speech (POS) tags. The entity definition component provides detailed definitions and identification criteria for each entity type in the dataset, including boundary delineation and clarification of classification confusion points. We design a selection procedure to choose few-shot examples that cover all entity types and those difficult entities to generate, implicitly specifying the output format. We sample the training data during the few-shot examples construction process, rather than using all the training data. The chain-of-thought component guides LLMs to provide reasoning for their output, enhancing the quality of generation. POS tags optionally supply syntactic information for contextual text. These components ensure that the prompts embody clear instructions, task-relevant background, and a defined output format, facilitating optimal model comprehension for the few-shot NER task.

GPT4NER differs from previous LLM-based NER methods in several aspects. Unlike PromptNER \cite{ashok2023promptnerpromptingnamedentity}, which randomly selects few-shot examples, GPT4NER implements a targeted selection strategy that ensures coverage of all entity types and difficult-to-generate entities, which improves reliability across multi-type scenarios. Compared to GPT-NER \cite{wang-etal-2025-gpt}, which queries one entity type at a time and requires a separate verification strategy to handle hallucinations, GPT4NER can query all entity types in a single call. Compared to BANER \cite{guo-etal-2025-baner}, which decomposes NER into a two-stage process and employs stage-specific prompt templates primarily focused on boundary detection, GPT4NER performs end-to-end entity recognition with prompts combining entity definitions, few-shot examples, chain-of-thought reasoning, and optional POS tags, allowing expressive instructions and structured output guidance.

To evaluate the effectiveness of GPT4NER, we conduct experiments on two benchmark datasets, CoNLL2003~\citep{sang2003introductionconll2003sharedtask} and OntoNotes5.0~\citep{pradhan-etal-2013-towards}, focusing on the few-shot NER task and its sub-task of named entity extraction (NEE). We compare the results of GPT4NER to two types of representative state-of-the-art models: few-shot models and fully-supervised models. Experimental results demonstrate that GPT4NER achieves the $F_1$ of 83.15\% on CoNLL2003 and 70.37\% on OntoNotes5.0, significantly outperforming few-shot baselines by an average margin of 7 points. Compared to fully-supervised baselines, GPT4NER achieves 87.9\% of their best performance on CoNLL2003 and 76.4\% of their best performance on OntoNotes5.0. Furthermore, our experiments utilize the relaxed-match metric, which is widely used for evaluating time expression recognition and normalization~\citep{verhagen-etal-2007-semeval, verhagen-etal-2010-semeval, uzzaman-etal-2013-semeval, zhong-etal-2017-time, 10.1145/3178876.3185997, zhong2020extracting, zhong2021time}, to evaluate the performance of few-shot models. Our analysis indicates that while few-shot models may not precisely recognize the boundaries of named entities, they can identify portions of these entities. Additionally, our findings highlight the importance of reporting model performance in the sub-task of few-shot NEE to better analyze model capabilities in the few-shot NER task.

In summary, the main contributions of this paper are as follows:

\begin{itemize}
	\item We propose GPT4NER, a method that prompts LLMs for few-shot NER. GPT4NER constructs effective prompts using three key components: entity definition, few-shot examples, and chain-of-thought, along with one optional component, POS tags, and adopts a targeted selection strategy that ensures coverage of all entity types and difficult-to-generate entities.
	
	\item We conduct experiments on two benchmark datasets, and experimental results demonstrate that GPT4NER significantly outperforms representative state-of-the-art few-shot models and achieves approximately 82.15\% of the best performance of fully-supervised models, on average.
	
	\item Our experiments suggest that utilizing the relaxed-match metric to evaluate model performance can enhance our understanding of model capabilities, and that reporting NEE performance provides further insights into model capabilities in the NER task. 
\end{itemize}

\section{Methodology}\label{sec:method}

\begin{figure}[t]
	\centering
	\includegraphics[width=1.01\columnwidth]{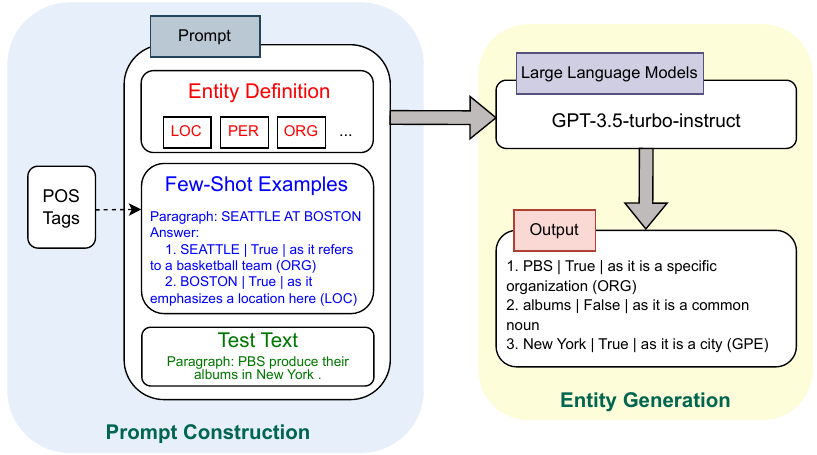}
	\caption{Overview of GPT4NER for few-shot NER. The left-hand side illustrates the prompt construction using three kinds of information: (1) entity definition, (2) few-shot examples with chain-of-thought reasoning, and (3) input test text. The right-hand side depicts the procedure of LLMs processing prompts and generating entities.}
	\label{fig:framework}
\end{figure}

Figure~\ref{fig:framework} provides an overview of GPT4NER for few-shot NER, comprising two parts: (1) prompt construction and (2) entity generation by LLMs. The prompt is built from three core elements: entity definitions, few-shot examples with chain-of-thought reasoning, and the input test text.

\subsection{Prompt Construction}

In leveraging the capabilities of LLMs, constructing effective prompts is crucial, laying the foundation for subsequent model training and inference processes. An exemplary prompt should embody the following three key characteristics to ensure optimal model comprehension and performance: 

\begin{itemize}
	\item  \textbf{Clear Instructions}: An effective prompt must provide explicit and unambiguous instructions regarding the objectives and requirements of the task. Such clarity is essential to facilitate accurate understanding of the core content.
	\item \textbf{Task-Relevant Background}: An effective prompt should integrate task-relevant background knowledge, encompassing domain-specific expertise, entity attributes, or several examples.
	\item \textbf{Output Format}: An effective prompt should specify the output format, either explicitly or implicitly, because the output format directly impacts subsequent processing and evaluation. A chaotic or disorganized output format can pose significant challenges for processing and evaluation tasks.
\end{itemize}

\begin{figure}[t]
	\centering
	\includegraphics[width=1.01\columnwidth]{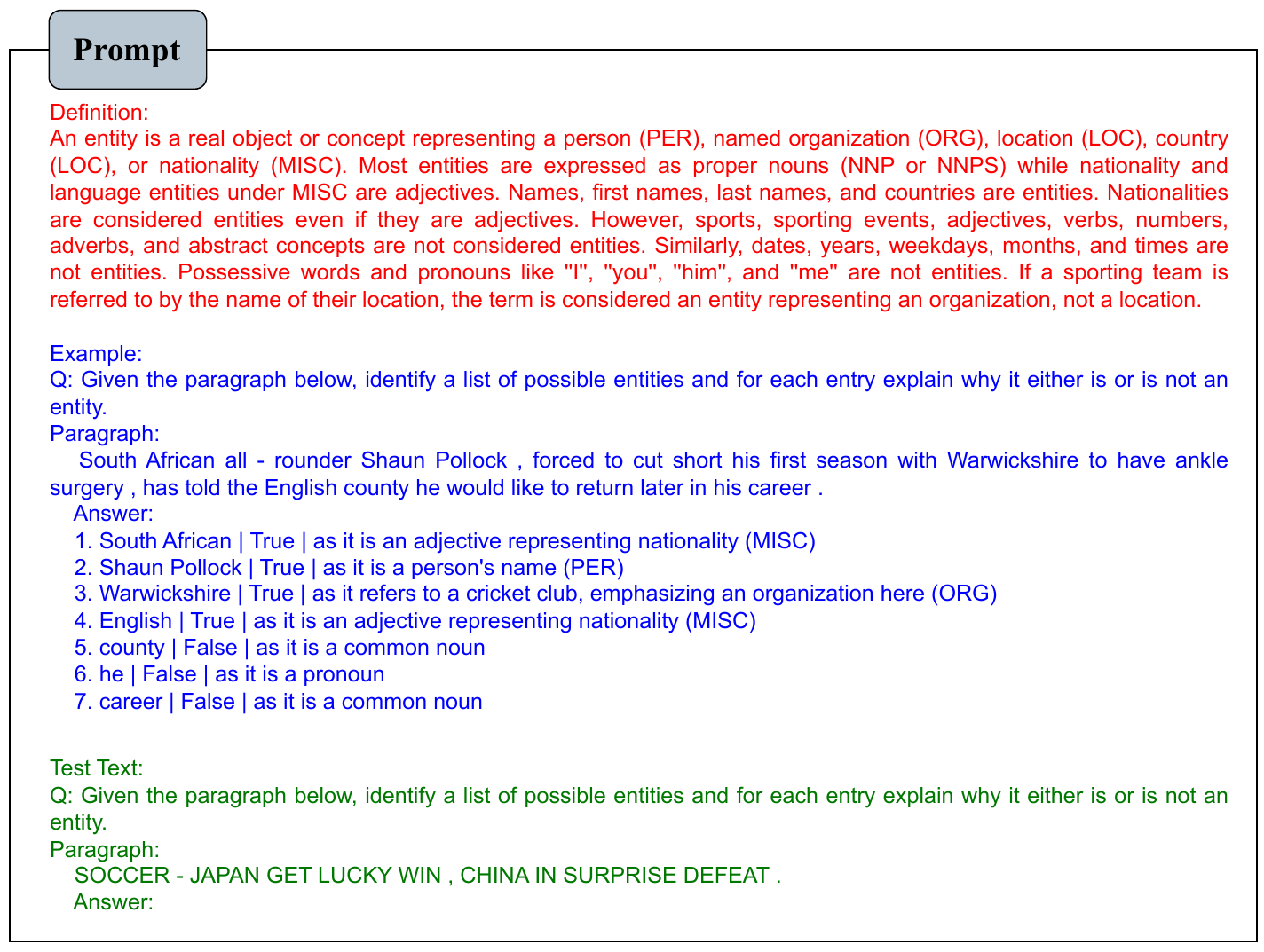}
	\caption{An example of effective prompts for the CoNLL2003 dataset. \textcolor{red}{Entity definition is in red}. \textcolor{blue}{Few-shot examples with question-answer format and chain-of-thought reason are in blue}. \textcolor[rgb]{0.25, 0.51, 0.2}{Test text is in dark green}.}
	
	
	\label{fig:prompt}
\end{figure}

To embody these key characteristics, we construct effective prompts that include three key components: (1) entity definition, (2) few-shot examples with output format, and (3) chain-of-thought, along with one optional component: syntactic POS information. Below, we detail these components with an example of effective prompts designed for the CoNLL2003 dataset, as illustrated in Figure~\ref{fig:prompt}.

\subsubsection{Entity Definition}

In different datasets, researchers may specify significantly diverse definitions and identification rules for the same types of entities. For example, both CoNLL2003 and OntoNotes5.0 include \ner{LOC} entities, but their definitions differ. CoNLL2003 classifies \ner{LOC} entities as countries, cities, regions such as ``London'' and ``Germany''. By contrast, OntoNotes5.0 defines \ner{LOC} entities as non-\ner{GPE} locations, including mountain ranges and planets. Furthermore, OntoNotes5.0 includes \ner{GPE} (i.e., geopolitical entities like countries and cities) and \ner{FAC} (i.e., facilities like buildings and roads), which can overlap with \ner{LOC} in some cases.

Therefore, simply hinting at differences between entity types in few-shot examples may impede the performance of traditional few-shot methods. It is crucial to provide explicit and unambiguous definitions and meticulous identification rules for each entity type within the dataset. Additionally, it is important to delineate boundary conditions and elucidate points of potential classification ambiguity. For example, OntoNotes5.0 specifies \ner{PERSON} to include generational markers (e.g., ``Jr.'' and ``IV'') while exclude honorifics (e.g., ``Ms.'' and ``Dr.'') and occupational titles (e.g., ``President'' and ``Secretary''). Explicitly describing the scope of an entity helps precisely identify the boundaries of entities.

An inherent challenge in prompt construction is reconciling the need for domain-specific knowledge with users' limited understanding. To resolve this challenge, we express entity definitions in natural language, avoiding excessive technical jargon. Such strategy not only facilitates comprehension but also provides greater flexibility, allowing for adaptable use across diverse datasets without sacrificing specificity.

In entity definition module, we adhere to these principles by providing detailed definitions and identification criteria in natural language for each entity type within individual datasets. These definitions include boundary delineation and points of classification confusion. In this paper, we utilize two benchmark datasets: CoNLL2003~\citep{sang2003introductionconll2003sharedtask} and OntoNotes5.0~\citep{pradhan-etal-2013-towards}. For CoNLL2003, we construct the following definition for its entities, as illustrated in Figure~\ref{fig:prompt}. For OntoNotes5.0, we construct the following definition for its entities:
\begin{quote}\label{qt:ontonotes}
	\textit{An entity is a real object or concept that represents an event, facility, country, language, location, nationality, organization, person, product, or work of art. Typically, entities are expressed as proper nouns (\ner{NNP} or \ner{NNPs}). Event (\ner{EVENT}) entities refer to proper nouns representing hurricanes, battles, wars, sports events, and attacks. Facility (\ner{FAC}) entities refer to proper nouns associated with man-made structures like buildings, airports, highways, and bridges. Geographical (\ner{GPE}) entities refer to proper nouns representing countries, cities, states, provinces, and municipalities. Language (\ner{LANGUAGE}) entities refer to named languages. Location (\ner{LOC}) entities refer to proper nouns representing non-\ner{GPE} locations, including mountain ranges, planets, geo-coordinates, bodies of water, named regions, and continents. Nationalities, religious, or political groups (\ner{NORP}) are expressed through adjectival forms of geographical, social, and political entities, location names, named religions, heritage, and political affiliations. Organization (\ner{ORG}) entities refer to proper nouns representing companies, government agencies, educational institutions, and sports teams. This also include adjectival forms of organization names and metonymic mentions of associated buildings or locations. Person (\ner{PERSON}) entities are represented by proper personal names, including fictional characters, first names, last names, nicknames, and generational markers (such as Jr. and IV), excluding occupational titles and honorifics. Product (\ner{PRODUCT}) entities refer to proper nouns representing model names, vehicles, or weapons. Manufacturer and product should be marked separately. Works of art (\ner{WORK\_OF\_ART}) refer to titles of books, songs, articles, television programs, or awards. If an organization, occupation title, and person's name form a phrase, then the organization and person's name is marked separately. Nominals and common nouns are not considered entities. Additionally, pronouns and pronominal elements are excluded from entities, as are contact information, plants, dates, years, times, numbers, legal documents, treaties, credit cards, checking accounts, \ner{CDs}, credit plans, financial instruments, and abstract concepts.}
\end{quote}


\subsubsection{Few-Shot Examples with Implicit Output Format}\label{sssec:few-shot-examples}

Few-shot examples serve as a vital instructional tool in prompts, providing tangible exemplars of contextual instantiation for each entity type. The inclusion of these examples aims to afford the model invaluable insights into contextual nuances underpinning entity identification.

Many LLMs have strict limitations on the maximum number of tokens they can process (e.g., OpenAI's GPT-3.5-turbo-instruct model supports only a 4K-token window). Consequently, each input can accommodate up to 10 examples, with around 400-500 tokens reserved for output.

In our constructed prompts, each example comprises three types of information: (1) \textit{a task-description question}, (2) \textit{an input text}, and (3) \textit{output results}. These few-shot examples provide direct instructions and evidence relevant to the task, enabling LLMs to grasp the logic of predictions.

\textbf{Task-Description Question}. The task-description question serves to guide LLMs on the task at hand. We utilize the following format as the task instruction:
\begin{itemize}
	\item \textit{Q: Given the paragraph below, identify a list of possible entities and for each entry explain why it either is or is not an entity.}
\end{itemize}

\textbf{Input Text.} The input text is selected as an example from the training data, with the primary objective of enhancing the accuracy of recognizing entities in test text. The selection process prioritizes texts that closely resemble the test text, especially those presenting identification challenges. These chosen texts often exhibit more complex results and involve entity categories that are easily confused, including both positive and negative instances (i.e., examples of both entities and non-entities). 

\begin{algorithm}[t]
	\caption{Example selection for limited tokens}
	\label{alg:example-selection}
	\begin{algorithmic}[1]
		\STATE \textbf{Input:} Training set $\mathcal{D}$, maximum token limit $T$
		\STATE \textbf{Output:} Optimized few-shot examples $\mathcal{P}$
		
		\STATE $\mathcal{P} \gets \emptyset$
		
		\STATE \textbf{Step 1: Select Texts with Multiple Entity Types}
		\STATE Select texts $\mathcal{T}_1 \subseteq \mathcal{D}$ with at least 3 entity types
		\STATE ${\mathcal{P} \gets \mathcal{P} \cup \{ \text{Select 3-4 texts from } \mathcal{T}_1 \text{ covering all entity types} \}}$
		\STATE \textbf{Step 2: Identify and Test Confused Entities}
		\STATE Select confused entities $\mathcal{E}_c \subseteq \mathcal{D}$
		\FOR {each text $t \in \mathcal{E}_c$}
		\STATE Test $t$ using current prompt $\mathcal{P}$
		\IF {entity recognition is suboptimal}
		\STATE $\mathcal{P} \gets \mathcal{P} \cup \{ t \}$
		\ENDIF
		\ENDFOR
		\STATE \textbf{Step 3: Sample and Finalize Examples}
		\WHILE {number of examples in $\mathcal{P}$ is less than 10 and token count $< T$}
		\STATE Randomly sample 10 texts $\mathcal{T}_2 \subseteq \mathcal{D}$ each time
		\FOR {each text $t \in \mathcal{T}_2$}
		\STATE Test $t$ using current prompt $\mathcal{P}$
		\IF {entity recognition is suboptimal}
		\STATE $\mathcal{P} \gets \mathcal{P} \cup \{ t \}$
		\ENDIF
		\ENDFOR
		\ENDWHILE
		
		\STATE \textbf{return} $\mathcal{P}$
	\end{algorithmic}
\end{algorithm}

Few-shot examples are thoughtfully selected to illustrate diverse contexts in which a given entity type may manifest, encompassing variations in syntactic structure and semantic context. By exposing the model to a range of context understanding and entity recognition instances, we aim to imbue the model with a robust understanding of the myriad manifestations of entity types, thereby enhancing its adaptability and generalization capabilities. To achieve this, we select and adjust examples through multiple sampling tests based on feedback from results. These few-shot examples include challenges in identifying entities and understanding specific contexts. Given the limited number of tokens specified by LLMs, we carefully select these examples. This selection procedure mainly comprises the following three steps, as illustrated in Algorithm~\ref{alg:example-selection}.

\begin{itemize}
	\item \textbf{Step 1:} \textbf{Select texts with multiple types of entities}. From the training set, select texts containing at least three types of entities. Choose three to four texts to ensure all entity types are covered for a 1-shot setup.
	\item \textbf{Step 2:} \textbf{Identify confused entities}. Conduct a small-scale test using the texts selected in \textbf{Step 1} to evaluate the prompt's effectiveness, and then add those texts whose entities are poorly generated as new examples to the prompt.
	\item \textbf{Step 3:} \textbf{Check for omissions}. Randomly select ten texts for testing each time. Gradually add examples following \textbf{Step 2} until the number of added examples reaches ten, the maximum number of examples.
\end{itemize}

\textbf{Output Format.} The output format is implicitly incorporated into prompts alongside entity definition and few-shot examples. It delineates the expected format for the output labels corresponding to identified entities. These implicit output formats serve as guiding beacons, steering the model towards generating output labels that adhere to predefined standards of clarity, consistency, and conciseness. By embedding an intrinsic awareness of annotation conventions within the model, these implicit output formats ensure that outputs are semantically accurate and adhere to established annotation standards.

Each test sentence needs to satisfy the following conditions: (1) it needs to clearly list words or phrases that are (or are not) entities and their corresponding category labels; (2) it needs to be easy for LLMs to learn and imitate, so that we can smoothly label each token in the test sentence. The output includes a list of candidate entities, explanations for identification and classification, and specific entity types. The output format is structured as follows:
\begin{quote}
	\textit{Candidate $|$ True or False $|$ Explanation of why the candidate is or is not an entity [(Type)] }
\end{quote}
which contains three elements:
\begin{itemize}
	\item \textit{Candidate}: This element indicates a generated candidate that may be considered as an entity.
	\item \textit{True or False}: This element indicates whether the generated candidate is an entity. Specifically, ``\textit{True}'' indicates that the candidate is an entity, while ``\textit{False}'' indicates that it is not.
	\item \textit{Explanation of why the candidate is or is not an entity [(Type)]}: This element explains why the candidate is or is not treated as an entity and specifies the type of the entity if the candidate is an entity. It is our designed chain-of-thought component and will be described in subsequent parts.
\end{itemize}

For example, as shown in Figure~\ref{fig:prompt}, the first entry of the output is ``\textit{South African $|$ True $|$ as it is an adjective representing nationality (MISC)}''. This means that ``\textit{South African}'' is a generated candidate that may be treated as an entity. ``\textit{True}'' indicates that ``\textit{South African}'' is indeed treated as an entity. The explanation ``\textit{as it is an adjective representing nationality (MISC)}'' clarifies why ``\textit{South African}'' is treated as an entity, specifically under the type \ner{MISC}. By contrast, the fifth entry, ``\textit{county $|$ False $|$ as it is a common noun}'', indicates that ``\textit{county}'' is a candidate, but ``\textit{False}'' suggests that this candidate is not an entity. The explanation, ``\textit{as it is a common noun}'', clarifies why ``\textit{county}'' is not treated as an entity.

\subsubsection{Chain-of-Thought}\label{ssec:CoT}

Incorporating explanations of whether candidates are entities into the prompt enhances the clarity of the instruction and serves as a practical implementation of our chain-of-thought reasoning. This approach strategically guides LLMs through a systematic thought process, encouraging careful consideration of each step and coherent explanations for its decisions. Recent findings suggest that chain-of-thought prompting can guide LLMs to output reasoning, even by simply adding ``think step by step'' to the prompt~\citep{10.5555/3600270.3602070,zhang2022automaticchainthoughtprompting, wang2022self}. Numerous studies have underscored the effectiveness of this approach, showing that guiding LLMs to think step by step and articulate their reasoning can greatly reduce errors. By asking LLMs to provide a reason for recognition along with generating the entity list, we can substantially improve the reliability of the outputs. In this work, chain-of-thought explanations are incorporated into the output format of the prompt.

\subsubsection{Part-of-Speech (POS) Information}\label{sssec:pos-tags}

\citet{zhong2020extracting} demonstrate that named entities are primarily composed of proper nouns, and \citet{ye2024llmdadataaugmentationlarge} show that data augmentation using LLMs to alter the syntactic structure of input text can enhance few-shot NER. Therefore, we incorporate part-of-speech (POS) tags to enrich the syntactic information of named entities in the text. Specifically, POS tags are included in these few-shot examples as part of the input text, as shown below:
\begin{quote}
	South/NNP African/JJ all/DT -/HYPH rounder/NN Shaun/NNP Pollock/NNP ,/, forced/VBN to/TO cut/VB short/IN his/PRP first/JJ season/NN with/IN Warwickshire/NNP to/TO have/VB ankle/NN surgery/NN ,/, has/VBZ told/VBN the/DT English/JJ county/NN he/PRP would/MD like/VB to/TO return/VB later/RBR in/IN his/PRP career/NN ./.
\end{quote}

\subsection{Entity Generation by Large Language Models}\label{ssec:entity-generation}

Recent studies approach the NER task as a sequence-to-sequence problem and employ methods such as prompt-based techniques or in-context learning. We adopt a similar perspective to address the NER task through sequence generation by prompting LLMs. A primary motivation for treating NER as sequence-generation problem is to mitigate the challenge of combinatorial explosion, which arises when entities consist of multiple tokens. Traditional token-based approaches may struggle to handle such cases effectively, leading to suboptimal performance and decreased accuracy. By contrast, LLMs have demonstrated noteworthy performance in NER tasks, even when they are trained on only a small subset of training data. This highlights the efficacy of leveraging unsupervised pre-trained models for sequence generation tasks, where the model can generalize effectively from a limited number of examples to achieve competitive performance across diverse datasets and domains. Unlike conventional supervised models that rely heavily on labeled training data, we utilize LLMs as a powerful sequence-generation tool for few-shot NER, capitalizing on their ability to perform well with minimal labeled data.

\subsection{Experimental Setup}\label{ssec:setup}

\subsubsection{Datasets.}\label{sssec:dataset}

The evaluation of GPT4NER is conducted on two benchmark datasets: CoNLL2003 \citep{sang2003introductionconll2003sharedtask} and OntoNotes5.0~\citep{pradhan-etal-2013-towards}. 

\textbf{CoNLL2003} is a widely used benchmark dataset derived from the Reuters RCV1 corpus, containing 1,393 news articles spanning from August 1996 to August 1997. It includes 35,089 entities categorized into four types: \ner{PER}, \ner{LOC}, \ner{ORG}, and \ner{MISC}.

\textbf{Ontonotes5.0} is also a widely used benchmark dataset developed for the analysis of several linguistic tasks in three languages. In this paper, we focus only the NER task in English and use only the NER portion of the OntoNotes5.0 dataset. This subset consists of 3,370 articles collected from various sources such as newswire and web data. It contains 18 types of entities, among which 10 types are primarily related to proper nouns or nationalities, while the other 8 types involve changing digits. We are mainly concerned with the 10 types of concrete entities related to proper nouns or nationalities: \ner{EVENT}, \ner{FAC}, \ner{GPE}, \ner{LANGUAGE}, \ner{LOC}, \ner{NORP}, \ner{ORG}, \ner{PERSON}, \ner{PRODUCT}, \ner{WORK\_OF\_ART}.\ftext{The excluded entity types are \ner{CARDINAL}, \ner{DATE}, \ner{LAW}, \ner{MONEY}, \ner{ORDINAL}, \ner{PERCENT}, \ner{QUANTITY}, and \ner{TIME}.}

For CoNLL2003, we follow previous studies~\citep{ma-hovy-2016-end} to divide its data into training, development, and testing sets. For OntoNotes5.0, we split the data into training, development, and testing sets using the same method as \citet{pradhan-etal-2013-towards}. Table~\ref{tb:data-statistics} summarizes the statistics of the two datasets.

\begin{table}
	\small
	\centering
	\caption{Statistics of the two benchmark datasets}
	\label{tb:data-statistics}
	\setlength{\tabcolsep}{1mm} 
	\begin{tabular}{@{}ccrrrc@{}}
		\toprule
		\multirow{1}{*}{\textbf{Dataset}}	& {} &	\multirow{1}{*}{\textbf{\#Sentences}} &	\multirow{1}{*}{\textbf{\#Tokens}} &	\multirow{1}{*}{\textbf{\#Entities}} &	\multirow{1}{*}{\textbf{\#Types}}\\	
		\midrule
		\multirow{3}{*}{CoNLL2003}	
		&Train & 14987 & 203621 & 23499 & \multirow{3}{*}{4} \\
		&Dev. & 3466 & 51362 & 5942 & \\
		&Test & 3684 & 46435 & 5648  &   \\
		\midrule
		\multirow{3}{*}{OntoNotes5.0}	
		&Train & 59924 & 1088503 & 55530  & \multirow{3}{*}{10}\\
		&Dev. & 8528 & 147724 & 7584  & \\
		&Test & 8262 & 152728 & 7505  & \\
		\bottomrule
	\end{tabular}
\end{table}

\subsubsection{State-of-the-Art Baselines}

We compare the performance of GPT4NER to five representative state-of-the-art models, including three few-shot models and two fully-supervised models.

\textbf{Few-shot baselines}:

\begin{itemize}
	\item \textbf{ProML}~\citep{chen-etal-2023-prompt} designs multiple prompt schemas to improve label semantics and introduces a novel architecture to combine these prompt-based representations. It targets tasks such as token set expansion and domain transfer.
	\item \textbf{CONTaiNER}~\citep{das-etal-2022-container} is a contrastive learning technique for few-shot NER that optimizes inter-token distribution distance using Gaussian-distributed embeddings. This method enhances differentiation between token categories and alleviates overfitting from training domains.
	\item \textbf{PromptNER}~\citep{ashok2023promptnerpromptingnamedentity} advances entity recognition by integrating entity definitions in addition to few-shot examples and prompts language models to produce a list of potential entities along with corresponding explanations.
\end{itemize}

\textbf{Fully-supervised baselines}:
\begin{itemize}
	\item \textbf{MRC-NER+DSC}~\citep{li-etal-2020-dice} employs dice loss instead of the standard cross-entropy objective for data-imbalanced NLP tasks. It uses a dynamic weight adjustment strategy that modifies training example weights, emphasizing hard-negative examples and reducing the impact of easy-negative ones. This model achieves state-of-the-art results on the OntoNotes5.0 dataset.
	\item \textbf{ACE+document-context}~\citep{wang-etal-2021-automated} utilizes reinforcement learning-based optimization with a novel reward function to automatically find the optimal combination of embeddings for structure prediction tasks. This model achieves state-of-the-art results on CoNLL2003.
\end{itemize}

\subsubsection{Evaluation Metrics} 

Like previous studies~\citep{wang-etal-2025-gpt,zhong2020extracting,li-etal-2020-dice}, we report the evaluation performance of each model using three standard metrics: \textit{Precision} ($Pre.$), \textit{Recall} ($Rec.$), and $F_1$, under both \textit{strict match} and \textit{relaxed match}.
\begin{equation}\label{eq:precision}
	Pre.=\frac{TP}{TP+FP}
\end{equation}
\begin{equation}\label{eq:recall}
	Rec.=\frac{TP}{TP+FN}
\end{equation}
\begin{equation}\label{eq:f1}
	F_1=\frac{2\times Pre.\times Rec.}{Pre. + Rec.}
\end{equation}
where $TP$ (true-positive) denotes the number of targets that appear in both the ground-truth and the prediction, $FP$ (false-positive) denotes the number of targets that are in the prediction but not in the ground-truth, while $FN$ (false-negative) denotes the number of targets that appear in the ground-truth but not appear in prediction.

\textit{Strict match} refers to an exact match between the recognized entities and the ground-truth entities, while \textit{relaxed match}~\citep{verhagen-etal-2007-semeval, verhagen-etal-2010-semeval, uzzaman-etal-2013-semeval, zhong-etal-2017-time, 10.1145/3178876.3185997, zhong2020extracting, zhong2021time} allows for some overlap between the recognized entities and the ground-truth entities.

\subsubsection{Implementation Details}

We use the GPT-3.5 (gpt-3.5-turbo-instruct) model as our LLMs backbone for all our experiments. This model supports a 4K-token context window. To maximize the utility of this capacity, we set the maximum output length to 400 tokens. Additionally, we set the temperature parameter to 0 so as to ensure reproducibility.

In addition, we use an open-source LLM, Llama3-8B~\citep{grattafiori2024llama} to compare. Also, we set the temperature parameter to 0 and the maximum output length to 400 tokens.

All our experiments are conducted on a server equipped with two Intel Xeon Gold 6240R CPUs (2.40GHz, 24 cores), 251GB of memory, and two NVIDIA RTX A5000 GPUs (24GB VRAM), running CentOS Linux 7 (Core). The server environment includes CUDA 12.1 and Python 3.7.5.

\begin{table*}[t]
	\centering
	\caption{Overall performance of GPT4NER and baselines in \textbf{named entity recognition (NER)}. Within each type of methods, the best results are in bold and the second best are underlined. Results marked with * indicate our reproduction. Results of ProML and CONTaiNER on Ontonotes5.0 are reported based on the average of three splits.}
	\label{tb:overall-performance}
	\begin{tabular}{@{}c|c|cccccc@{}}
		\toprule
		\multirow{2}{*}{\textbf{Dataset}}	&	\multirow{2}{*}{\textbf{Method}}	&	\multicolumn{3}{c}{\textbf{Strict Match}}	&	\multicolumn{3}{c}{\textbf{Relaxed Match}}	\\
		\cmidrule(lr){3-5}\cmidrule(lr){6-8}
		&			&	$Pre.$	&	$Rec.$	&	$F_1$	&	$Pre.$	&	$Rec.$	&	$F_1$	\\	
		\midrule
		\multirow{15}{*}{CoNLL2003}		
		&BERT\_MRC+DSC 	&\textbf{93.41} & \textbf{93.25} & \underline{93.33} &- &- &-			\\
		&ACE+document-context	&- &- &	\textbf{94.60} &- &- &-		\\
		\cmidrule(lr){2-8}
		&ProML(1shot) 			&- 		&- 		&69.16 	&- &- &-	\\
		&ProML(5shot) 			&- 		&- 		&79.16 	&- &- &-           \\
		&CONTaiNER(1shot) 	&- 		&- 		&57.80 	&- &- &-		\\
		&CONTaiNER(5shot) 	&- 		&- 		&72.80 	&- &- &-		\\
		&PromptNER			&-		&-		&78.62	&-	&-	&-	\\
		&ProML(1shot)*		& 63.26 	& 65.05 	& 64.10 	& 76.79 	& 78.97 	& 77.81	\\
		&ProML(5shot)* 		& 77.60 	& 80.15 	& 78.84  	& \textbf{85.28} 	& 88.08 	& \textbf{86.65}	\\
		&CONTaiNER(1shot)*	& 61.47 	& 61.10 	& 61.27 	& 66.80 	& 66.42 	& 66.59 	\\
		&CONTaiNER(5shot)* 	& 72.42 	& 74.89 	& 73.62 	& 77.91 	& 81.58 	& 79.21	\\
		&PromptNER*			&66.68	&70.13	&68.36	&69.71	&73.32	&71.47	\\
		&GPT4NER-Llama3		& 67.85	&72.93	&70.30	&72.21	&77.62	&74.82	\\
		&\textbf{GPT4NER} (ours)		& \textbf{79.20} 	& \textbf{87.52} 	& \textbf{83.15} 	& \underline{81.56} & \textbf{90.12} & \underline{85.63}		\\
		&\textbf{GPT4NER} w/o POS		& \underline{78.24} 	& \underline{86.05} 	& \underline{81.96} 	& 80.96 & \underline{89.04} & 84.81		\\
		\midrule
		\multirow{12}{*}{OntoNotes5.0}	
		&BERT\_MRC+DSC &\textbf{91.59} &\textbf{92.56} & \textbf{92.07} &- &- &-		\\
		\cmidrule(lr){2-8}
		&ProML(1shot)			&- 		&- 		&45.98 	&- &- &-	\\
		&ProML(5shot)  		&- 		&- 		&63.24 	&- &- &-            \\
		&CONTaiNER(1shot)		&- 		&- 		&32.00 	&- &- &-		\\
		&CONTaiNER(5shot)		&- 		&- 		&56.20 	&- &- &-		\\
		&ProML(1shot)* 		&36.53 	&51.59 	&42.74 	& 52.42 	& 74.17 	& 61.39	\\
		&ProML(5shot)* 		&50.21 	&64.46 	&56.42 	& 65.91 	& 84.80 	& 74.13	\\
		&CONTaiNER(1shot)*	&40.92 	&33.61 	&36.84 	& 61.68 	& 50.30 	& 55.31	\\
		&CONTaiNER(5shot)*	&54.49 	&53.64 	&54.06 	& 73.47 	& 72.45 	& 72.95	\\
		&GPT4NER-Llama3		& 37.52	&55.63	&44.82	& 45.21	&67.02	&53.99	\\
		&\textbf{GPT4NER} (ours)	&\underline{62.66} 	&\underline{71.32} 	& \underline{66.71} 	&\underline{74.62} 	& \underline{84.93} & \underline{79.44}		\\
		&\textbf{GPT4NER} w/o POS	&\textbf{67.15} 	&\textbf{73.92} 	& \textbf{70.37} 	&\textbf{79.85} 	& \textbf{87.90} & \textbf{83.68}		\\
		\bottomrule
	\end{tabular}
\end{table*}

\section{Results and Discussion}\label{sec:experiments}

We evaluate the effectiveness of GPT4NER on two benchmark datasets, CoNLL2003 \citep{sang2003introductionconll2003sharedtask} and OntoNotes5.0~\citep{pradhan-etal-2013-towards}, against five representative state-of-the-art models. These include three few-shot models, namely ProML~\citep{chen-etal-2023-prompt}, CONTaiNER~\citep{das-etal-2022-container}, and PromptNER~\citep{ashok2023promptnerpromptingnamedentity}, and two fully-supervised models, MRC-NER+DSC~\citep{li-etal-2020-dice} and ACE+document-context~\citep{wang-etal-2021-automated}.

\subsection{Experimental Results}\label{ssec:results}

We present experimental results on two tasks: (1) named entity recognition (NER), which aims to extract named entities from free text and then categorize them into predefined types, and (2) named entity extraction (NEE), which is also known as entity boundary detection that aims to simply extract named entities from free text without classifying them into specific types.

\subsubsection{Experimental Results on Named Entity Recognition}\label{sssec:ner-results}

Table~\ref{tb:overall-performance} presents the overall performance of GPT4NER and the five baselines on the two benchmark datasets in the NER task. For the three few-shot baselines, we include results reported in their original papers as well as results reproduced in our study, marked with *. Compared to the three few-shot baselines, among the total 12 measures (i.e., 3 metrics $\times$ 2 match types $\times$ 2 datasets), GPT4NER achieves the best performance in 10 measures and second-best in 12 measures, except for $Pre.$ and $F_1$ under relaxed match on CoNLL2003. Specifically, GPT4NER achieves the $F_1$ of 83.15\% under strict match on CoNLL2003 and 70.37\% on OntoNotes5.0, surpassing few-shot baselines by at least 4.0 points and 7.1 points on the two datasets, respectively. Under relaxed match, GPT4NER achieves the $F_1$ of 83.68\% on OntoNotes5.0, outperforming few-shot baselines by at least 9.5 points. On CoNLL2003, GPT4NER achieves the $F_1$ of 85.63\%, which is slightly below the best result of few-shot baselines (i.e., 86.65\%). Compared to the two fully-supervised baselines, GPT4NER achieves 87.9\% of their best performance on CoNLL2003 and 76.4\% of their best performance on OntoNotes5.0 in terms of the $F_1$ under strict match. Compared to Llama3-8B model, GPT4NER outperforms at least 12.8 points under strict match and 10.8 points under relaxed match on CoNLL2003. On Ontonotes5.0, GPT4NER outperforms 25.5 points under strict match and 29.6 points under relaxed match than Llama3-8B.

\textbf{GPT4NER vs. Few-Shot Baselines.} Let's compare GPT4NER to few-shot baselines. Table~\ref{tb:overall-performance} illustrates that under strict match, GPT4NER significantly outperforms all three few-shot baselines across all three metrics on both datasets. Specifically, GPT4NER shows the $F_1$ improvement of at least 3.99 points on CoNLL2003 and at least 7.13 points on OntoNotes5.0. This demonstrates GPT4NER's superior ability to accurately generate named entities with predefined types. Under relaxed match, GPT4NER also outperforms all three few-shot baselines across all three metrics on both datasets, with the exception of ProML (5-shot) on CoNLL2003 in terms of $Pre.$ and $Rec.$. Notably, GPT4NER achieves the highest $Rec.$ on both datasets, indicating its strong capability to generate named entities with predefined types under lenient condition.

\textbf{GPT4NER vs. Fully-Supervised Baselines.} The two fully-supervised baselines achieve state-of-the-art performance on both CoNLL2003 and OntoNotes5.0 by leveraging large amounts of annotated training data. As shown in Table~\ref{tb:overall-performance}, GPT4NER trails behind the best performance of the fully-supervised baselines by 11.5 points on CoNLL2003 and by 21.7 points on OntoNotes5.0. However, fully-supervised baselines~\citep{wang-etal-2021-automated,li-etal-2020-dice} require extensive annotated training data and perform poorly with less training data. The performance of supervised models increase with the training data~\citep{wang-etal-2025-gpt}. By contrast, GPT4NER uses only a few labeled examples with minimal human effort in prompting LLMs, but still achieves 87.9\% of fully-supervised baselines' best performance on CoNLL2003 and 76.4\% on OntoNotes. This demonstrates the potential of GPT4NER for few-shot NER, especially in low-resource scenarios.

\textbf{Strict Match vs. Relaxed Match.} Table~\ref{tb:overall-performance} shows that all few-shot models perform better under relaxed match compared to strict match across all metrics and datasets. It shows that for all four models, the scores under relaxed match are significantly higher than the corresponding ones under strict match across all three metrics on both datasets. To illustrate the usefulness of utilizing relaxed match in addition to strict match for evaluating performance, we define a metric called ``\textbf{score improvement} ($SI$)'' as Eq.~(\ref{eq:si}) to denote the difference between the scores under relaxed match and under strict match that are achieved by a model on a dataset:
\begin{equation}\label{eq:si}
	SI(m)=Relaxed(m)-Strict(m)
\end{equation}
where $Relaxed$ denotes the score under relaxed match, $Strict$ denotes the score under strict match, and $m\in\{Pre., Rec., F_1\}$. For example, GPT4NER achieves the $SI(F_1)$ of 2.48 points (i.e., $2.48=85.63-83.15$) on CoNLL2003.

\begin{table}[t]
	\small
	\centering
	\caption{SI value of GPT4NER and baselines in \textbf{named entity recognition (NER)}. Within each type of methods, the smallest results are in bold and the second smallest are underlined. Results marked with * indicate our reproduction.}
	\label{tb:NER-SI}
	\begin{tabular}{@{}c|c|ccc@{}}
		\toprule
		\multirow{2}{*}{\textbf{Dataset}}	&	\multirow{2}{*}{\textbf{Method}}	&	\multicolumn{3}{c}{\textbf{SI}}\\
		&			&	$Pre.$	&	$Rec.$	&	$F_1$	\\	
		\midrule
		\multirow{8}{*}{CoNLL2003}		
		&ProML(1shot)*		& 13.53 & 13.92 & 13.71 		\\
		&ProML(5shot)* 		& 7.68 	& 7.93 	& 7.81  	\\
		&CONTaiNER(1shot)*	& 5.33 	& 5.32 	& 5.32 	\\
		&CONTaiNER(5shot)* 	& 5.49 	& 6.69 	& 5.59 	\\
		&PromptNER*			& 3.03	& 3.19	& 3.11	\\
		&GPT4NER-Llama3		& 4.36	& 4.69	& 4.52	\\
		&\textbf{GPT4NER} (ours)		& \textbf{2.36} 	& \textbf{2.60} 	& \textbf{2.48} 	\\
		&\textbf{GPT4NER} w/o POS		& \underline{2.72} 	& \underline{2.99} 	& \underline{2.85} 	\\
		\midrule
		\multirow{7}{*}{OntoNotes5.0}	
		&ProML(1shot)* 		&15.89 	&22.58 	&18.65 	\\
		&ProML(5shot)* 		&15.70 	&20.34 	&17.71 	\\
		&CONTaiNER(1shot)*	&20.76 	&16.69 	&18.47 	\\
		&CONTaiNER(5shot)*	&18.98 	&18.81 	&18.89 	\\
		&GPT4NER-Llama3		&\textbf{7.69}	&\textbf{11.39}	&\textbf{9.17}	\\
		&\textbf{GPT4NER} (ours)	&\underline{11.96} 	&\underline{13.61} 	& \underline{12.73} 	\\
		&\textbf{GPT4NER} w/o POS	&12.70 	&13.98 	&13.31 	\\
		\bottomrule
	\end{tabular}
\end{table}

As shown in Table~\ref{tb:NER-SI}, the four few-shot models achieve the $SI(Pre.)$ of 2.36$\sim$13.53 points, the $SI(Rec.)$ of 2.60$\sim$13.92 points, and the $SI(F_1)$ of 2.48$\sim$13.71 points on CoNLL2003. On OntoNotes5.0, the $SI(Pre.)$ values are 7.69$\sim$20.76 points, the $SI(Rec.)$ values are 11.39$\sim$22.58 points, and the $SI(F_1)$ values are 9.17$\sim$18.89 points. These high $SI(\cdot)$ values indicate that models may struggle to exactly recognize the boundaries of named entities but can partially recognize these named entities. Additionally, the $SI(Pre.)$, $SI(Rec.)$, and $SI(F_1)$ values on OntoNotes5.0 are significantly greater than the corresponding ones on CoNLL2003. This difference could be due to the more complex and diverse text in OntoNotes5.0, which includes more syntactic and semantic variations. The few-shot models find it challenging to accurately recognize or generate the precise boundaries of entities in such complex and diverse texts, leading to a noticeable performance difference between relaxed match and strict match.

\begin{table*}[t]
	\centering
	\caption{Performance of GPT4NER and few-shot baselines in \textbf{named entity extraction (NEE)}. The best results are highlighted in bold and the second best are underlined. Results marked with * indicate our reproduction.}
	\label{tb:nee-result}
	\begin{tabular}{@{}c|c|cccccc@{}}
		\toprule
		\multirow{2}{*}{\textbf{Dataset}}	&	\multirow{2}{*}{\textbf{Method}}	&	\multicolumn{3}{c}{\textbf{Strict Match}}	&	\multicolumn{3}{c}{\textbf{Relaxed Match}}	\\
		\cmidrule(lr){3-5}\cmidrule(lr){6-8}
		&			&	$Pre.$	&	$Rec.$	&	$F_1$	&	$Pre.$	&	$Rec.$	&	$F_1$	\\	
		\midrule
		\multirow{7}{*}{CoNLL2003}		
		&ProML(1shot)*			& 72.21 	& 74.21 	& 73.14 	& 89.10 & 91.60 & 90.27	\\
		&ProML(5shot)* 			& 83.09 	& 85.82 	& 84.42  	& \underline{92.83} & 95.88 & \textbf{94.32}           \\
		&CONTaiNER(1shot)*		& 82.20 	& 81.81 	& 81.98 	& \textbf{92.97} & 92.57 & 92.75 		\\
		&CONTaiNER(5shot)* 		& \underline{83.54} 	& 86.45 	& 84.96 & 92.38 & 95.60 &\underline{ 93.95}			\\
		&GPT4NER-Llama3		& 76.66	&82.38	&79.42	&83.97	&90.24	&86.99	\\
		&\textbf{GPT4NER} (ours)	& \textbf{83.93} 	& \textbf{92.74} 	& \textbf{88.12} & 88.13 & \textbf{97.38} & 92.52		\\
		&\textbf{GPT4NER} w/o POS	& 82.58 	& \underline{90.83} 	& \underline{86.51} & 87.69 & \underline{96.44} & 91.85		\\
		\midrule
		\multirow{7}{*}{OntoNotes5.0}	
		&ProML(1shot)* 			&38.43 	&54.32 	&44.98 & 57.07 & 80.88 & 66.87		\\
		&ProML(5shot)* 			&51.84 	&66.59 	&58.27 & 69.78 & 89.85 & 78.50        \\
		&CONTaiNER(1shot)*		&43.04 	&35.32 	&38.73 & 67.14 & 54.68 & 60.16		\\
		&CONTaiNER(5shot)*		&56.24 	&55.38 	&55.80 & 77.67 & 76.61 & 77.13		\\
		&GPT4NER-Llama3			& 41.27	&61.16	&49.28	&53.07	&78.65	&63.38	\\
		&\textbf{GPT4NER} (ours)	& \underline{66.67} 	& \underline{75.88} 	& \underline{70.98} 	& \underline{82.04} 	& \underline{93.38} & \underline{87.34}		\\
		&\textbf{GPT4NER} w/o POS	& \textbf{70.72} 	& \textbf{77.85} 	& \textbf{74.12} 	& \textbf{86.48} 	& \textbf{95.20} & \textbf{90.63}		\\
		\bottomrule
	\end{tabular}
\end{table*}

These high $SI(Pre.)$, $SI(Rec.)$, and $SI(F_1)$ values may be attributed to the models' recognition or generation capabilities. However, annotation inconsistencies could also be a contributing factor. For example, within the same dataset, some \ner{PER}/\ner{PERSON} entities may include prefix words (e.g., ``Mr.'' and ``Dr.''), while others may exclude these prefixes. Furthermore, some loose recognitions of entity boundaries are acceptable. 

As shown in the following two examples of GPT4NER on OntoNotes5.0, the model predicts ``Dick Cheney 's'' as a \ner{PERSON} and ``the Reporters ' Committee for Freedom of the Press'' as a \ner{ORG}. The two predictions are slightly different from the corresponding ground-truth, and under strict match, they are considered wrong. However, under relaxed match, they are considered correct. This demonstrates that relaxed match evaluates performance in a broader sense and provides a more comprehensive assessment of the model, which is closer to real-world applications. Therefore, we consider relaxed match a valuable metric, complementary to strict match, for evaluating model performance.

\begin{itemize}
	\item \textit{Test Text}: In a separate first person account Miller confirmed that she told the grand jury that Scooter Libby Dick Cheney 's top aide discussed with her as many as three times the role of Valerie Plame as a CIA employee .
	
	\textit{Gold label}: ``Miller'': ``\ner{PERSON}'', ``Scooter Libby'': ``\ner{PERSON}'', ``\textbf{Dick Cheney 's}'': ``\ner{PERSON}'', ``Valerie Plame'': ``\ner{PERSON}'', ``CIA'': ``\ner{ORG}''
	
	\textit{Prediction}: ``Miller'': ``\ner{PERSON}'', ``Scooter Libby'': ``\ner{PERSON}'', ``\textbf{Dick Cheney}'': ``\ner{PERSON}'', ``Valerie Plame'': ``\ner{PERSON}'', ``CIA'': ``\ner{ORG}''
	
	
	\item \textit{Test Text}: in Minneapolis Lucy Dalglish executive director of the Reporters ' Committee for Freedom of the Press .
	
	\textit{Gold label}: ``Minneapolis'': ``\ner{GPE}'', ``Lucy Dalglish'': ``\ner{PERSON}'', ``\textbf{the Reporters ' Committee for Freedom of the Press}'': ``\ner{ORG}''
	
	\textit{Prediction}: ``Minneapolis'': ``\ner{GPE}'', ``Lucy Dalglish'': ``\ner{PERSON}'', ``\textbf{Reporters ' Committee for Freedom of the Press}'': ``\ner{ORG}'
	
\end{itemize}

\subsubsection{Experimental Results on Named Entity Extraction}\label{sssec:nee-results}

Table~\ref{tb:nee-result} reports the overall performance of GPT4NER and the few-shot baselines on the two benchmark datasets in the NEE task. The results of the few-shot baselines are our reproductions, marked with an asterisk (*). Among the total 12 measures, GPT4NER achieves 10 best results and 9 second-best ones, except for $Pre.$ and $F_1$ under relaxed match and $Pre.$ under strict match on CoNLL2003. Specifically, GPT4NER attains the $F_1$ of 88.12\% under strict match on CoNLL2003 and 74.12\% on OntoNotes5.0, significantly outperforming the few-shot baselines by at least 3.1 points on CoNLL2003 and at least 15.8 points on OntoNotes5.0. Under relaxed match, GPT4NER achieves the $F_1$ of 90.63\% on OntoNotes5.0, surpassing the few-shot baselines by at least 12.1 points. On CoNLL2003, GPT4NER achieves the $F_1$ of 92.52\%, which is slightly lower than the best result of the few-shot baselines (i.e., 94.32\%). GPT4NER outperforms Llama3-8B by 8.7 points under strict match and 5.5 points under relaxed match on CoNLL2003. On Ontonotes5.0, GPT4NER outperforms Llama3-8B by 24.8 points under strict match and 27.2 points under relaxed match. These results are consistent with those reported in Table~\ref{tb:overall-performance}, confirming the effectiveness and robustness of GPT4NER in few-shot NER and its sub-task.

\begin{table}[t]
	\small
	\centering
	\caption{SI value of GPT4NER and baselines in \textbf{named entity extraction (NEE)}. Within each type of methods, the smallest results are in bold and the second smallest are underlined. Results marked with * indicate our reproduction.}
	\label{tb:NEE-SI}
	\begin{tabular}{@{}c|c|ccc@{}}
		\toprule
		\multirow{2}{*}{\textbf{Dataset}}	&	\multirow{2}{*}{\textbf{Method}}	&	\multicolumn{3}{c}{\textbf{SI}}\\
		&			&	$Pre.$	&	$Rec.$	&	$F_1$	\\	
		\midrule
		\multirow{7}{*}{CoNLL2003}		
		&ProML(1shot)*		& 16.89 	& 17.39 	& 17.13	\\
		&ProML(5shot)* 		& 9.74 	& 10.06 	& 9.90  	\\
		&CONTaiNER(1shot)*	& 10.77 	& 10.76 	& 10.77 	\\
		&CONTaiNER(5shot)* 	& 8.84 	& 9.15 	& 8.99 	\\
		&GPT4NER-Llama3	    & 7.31	& 7.86	& 7.57	\\
		&\textbf{GPT4NER} (ours)		& \textbf{4.20} 	& \textbf{4.64} 	& \textbf{4.40} 	\\
		&\textbf{GPT4NER} w/o POS		& \underline{5.11} 	& \underline{5.61} 	& \underline{5.34} 	\\
		\midrule
		\multirow{7}{*}{OntoNotes5.0}	
		&ProML(1shot)* 		&18.64 	&26.56 	&21.89 	\\
		&ProML(5shot)* 		&17.94 	&23.26 	&20.23 	\\
		&CONTaiNER(1shot)*	&24.10 	&19.36 	&21.43 	\\
		&CONTaiNER(5shot)*	&21.43 	&21.23 	&21.33 	\\
		&GPT4NER-Llama3		&\textbf{11.80}	&\underline{17.49}	&\textbf{14.10}	\\
		&\textbf{GPT4NER} (ours)	&\underline{15.37} 	&17.50 	& \underline{16.36} 	\\
		&\textbf{GPT4NER} w/o POS	&15.76 	&\textbf{17.35} 	&16.51 	\\
		\bottomrule
	\end{tabular}
\end{table}

\begin{table*}[t]
	\centering
	\caption{Detailed comparison of GPT4NER and its Chain-of-Thought ablation in \textbf{named entity recognition (NER)}, by entity type, including precision, recall, $F_1$ under strict match, and counts. Specifically, GPT4NER with POS tags serves as the baseline for CoNLL2003, while GPT4NER without POS tags serves as the baseline for OntoNotes5.0.}
	\label{tb:ner-type-detail-result}
	\begin{tabular}{@{}c|c|c|ccc|ccc@{}}
		\toprule
		\multirow{2}{*}{\textbf{Dataset}} & \multirow{2}{*}{\textbf{Method}} & \multirow{2}{*}{\textbf{Entity Type}} & 
		\multicolumn{3}{c|}{\textbf{Strict Match}} & \multicolumn{3}{c}{\textbf{Entity Count}} \\
		\cmidrule(lr){4-6} \cmidrule(lr){7-9} 
		& & & \textbf{Pre.} & \textbf{Rec.} & \textbf{F1} & \textbf{Gold} & \textbf{Pred} & \textbf{Correct} \\	
		\midrule
		\multirow{8}{*}{CoNLL2003} 
		& \multirow{4}{*}{\textbf{GPT4NER}} 
		& PER    & 93.52 & 94.56 & 94.03 & 1617 & 1635 & 1529 \\
		& & LOC    & 88.44 & 90.35 & 89.38 & 1668 & 1704 & 1507 \\
		& & ORG    & 69.97 & 87.66 & 77.82 & 1661 & 2081 & 1456 \\
		& & MISC   & 55.34 & 64.25 & 59.46 & 702  & 815  & 451  \\
		\cmidrule(lr){2-9}
		& \multirow{4}{*}{w/o Chain-of-thought} 
		& PER    & 94.77 & 94.12 & 94.45 & 1617 & 1606 & 1522 \\
		& & LOC    & 70.53 & 94.96 & 80.94 & 1668 & 2246 & 1584 \\
		& & ORG    & 68.83 & 64.48 & 66.58 & 1661 & 1556 & 1071 \\
		& & MISC   & 63.98 & 63.25 & 63.61 & 702  & 694  & 444  \\
		\midrule
		\multirow{20}{*}{OntoNotes5.0} 
		& \multirow{10}{*}{\textbf{GPT4NER}} 
		& PERSON      & 76.20 & 74.90 & 75.55 & 1988 & 1954 & 1489 \\
		& & ORG        & 64.57 & 64.18 & 64.38 & 1795 & 1784 & 1152 \\
		& & LOC        & 21.46 & 59.22 & 31.50 & 179  & 494  & 106  \\
		& & NORP       & 64.79 & 78.12 & 70.84 & 841  & 1014 & 657  \\
		& & GPE        & 90.82 & 84.33 & 87.45 & 2240 & 2080 & 1889 \\
		& & FAC        & 28.46 & 27.41 & 27.92 & 135  & 130  & 37   \\
		& & EVENT      & 20.31 & 41.27 & 27.23 & 63   & 128  & 26   \\
		& & PRODUCT    & 17.93 & 68.42 & 28.42 & 76   & 290  & 52   \\
		& & LANGUAGE   & 55.56 & 68.18 & 61.22 & 22   & 27   & 15   \\
		& & WORK\_OF\_ART& 36.44 & 75.30 & 49.12 & 166  & 343  & 125  \\
		\cmidrule(lr){2-9}
		& \multirow{10}{*}{w/o Chain-of-thought} 
		& PERSON      & 80.04 & 77.87 & 78.94 & 1988 & 1934 & 1548 \\
		& & ORG        & 65.60 & 65.13 & 65.36 & 1795 & 1782 & 1170 \\
		& & LOC        & 25.60 & 59.22 & 35.75 & 179  & 414  & 106  \\
		& & NORP       & 75.34 & 78.83 & 77.05 & 841  & 880  & 663  \\
		& & GPE        & 91.21 & 84.82 & 87.90 & 2240 & 2083 & 1900 \\
		& & FAC        & 26.88 & 37.04 & 31.15 & 135  & 186  & 50   \\
		& & EVENT      & 19.61 & 31.75 & 24.24 & 63   & 102  & 20   \\
		& & PRODUCT    & 29.71 & 68.42 & 41.43 & 76   & 175  & 52   \\
		& & LANGUAGE   & 62.96 & 77.27 & 69.39 & 22   & 27   & 17   \\
		& & WORK\_OF\_ART& 47.79 & 71.69 & 57.35 & 166  & 249  & 119  \\
		\bottomrule
	\end{tabular}
\end{table*}

\textbf{Strict Match vs. Relaxed Match.} We utilize the score improvement ($SI$) as defined by Eq.~(\ref{eq:si}) to illustrate model performance in the NEE task. Table~\ref{tb:NEE-SI} shows that the three few-shot models achieve $SI(Pre.)$ values of 4.20$\sim$16.89 points, $SI(Rec.)$ values of 4.64$\sim$17.39 points, and $SI(F_1)$ values of 4.40$\sim$17.13 points on CoNLL2003. On OntoNotes5.0, the few-shot models achieve $SI(Pre.)$ values of 11.80$\sim$24.10 points, $SI(Rec.)$ values of 17.35$\sim$26.56 points, and $SI(F_1)$ values of 14.10$\sim$21.89 points. These $SI(Pre.)$, $SI(Rec.)$, and $SI(F_1)$ values are consistent with those in the NER task.\ftext{In fact, the $SI(Pre.)$, $SI(Rec.)$, and $SI(F_1)$ values in NEE are even higher than those in NER.} These high $SI(Pre.)$, $SI(Rec.)$, and $SI(F_1)$ values confirm the usefulness and necessity of utilizing relaxed match as a complement to evaluate model performance in NER and its sub-task.

\begin{table*}[t]
	\centering
	\caption{Detailed comparison of GPT4NER and its Chain-of-Thought ablation in \textbf{named entity extraction (NEE)}, by entity type, including precision, recall, $F_1$ under strict match, and counts. Specifically, GPT4NER with POS tags serves as the baseline for CoNLL2003, while GPT4NER without POS tags serves as the baseline for OntoNotes5.0.}
	\label{tb:nee-type-detail-result}
	\begin{tabular}{@{}c|c|c|ccc|ccc@{}}
		\toprule
		\multirow{2}{*}{\textbf{Dataset}} & \multirow{2}{*}{\textbf{Method}} & \multirow{2}{*}{\textbf{Entity Type}} & 
		\multicolumn{3}{c|}{\textbf{Strict Match}} & \multicolumn{3}{c}{\textbf{Entity Count}} \\
		\cmidrule(lr){4-6} \cmidrule(lr){7-9} 
		& & & \textbf{Pre.} & \textbf{Rec.} & \textbf{F1} & \textbf{Gold} & \textbf{Pred} & \textbf{Correct} \\	
		\midrule
		\multirow{8}{*}{CoNLL2003}
		& \multirow{4}{*}{\textbf{GPT4NER}}
		& PER    & 94.74 & 95.79 & 95.26 & 1617 & 1635 & 1549 \\
		& & LOC    & 94.95 & 97.00 & 95.97 & 1668 & 1704 & 1618 \\  
		& & ORG    & 72.32 & 90.61 & 80.44 & 1661 & 2081 & 1505 \\ 
		& & MISC   & 69.45 & 80.63 & 74.62 & 702  & 815  & 566  \\  
		\cmidrule(lr){2-9}
		& \multirow{4}{*}{w/o Chain-of-thought}  
		& PER    & 96.45 & 95.79 & 96.12 & 1617 & 1606 & 1549 \\  
		& & LOC    & 71.86 & 96.76 & 82.47 & 1668 & 2246 & 1614 \\ 
		& & ORG    & 99.16 & 92.90 & 95.93 & 1661 & 1556 & 1543 \\  
		& & MISC   & 79.97 & 79.06 & 79.51 & 702  & 694  & 555  \\  
		\midrule
		\multirow{20}{*}{OntoNotes5.0}  
		& \multirow{10}{*}{\textbf{GPT4NER}}  
		& PERSON      & 77.84 & 76.51 & 77.17 & 1988 & 1954 & 1521 \\  
		& & ORG        & 69.67 & 69.25 & 69.46 & 1795 & 1784 & 1243 \\  
		& & LOC        & 22.87 & 63.13 & 33.58 & 179  & 494  & 113  \\  
		& & NORP       & 66.77 & 80.50 & 72.99 & 841  & 1014 & 677  \\ 
		& & GPE        & 95.14 & 88.35 & 91.62 & 2240 & 2080 & 1979 \\  
		& & FAC        & 66.92 & 64.44 & 65.66 & 135  & 130  & 87   \\  
		& & EVENT      & 20.31 & 41.27 & 27.23 & 63   & 128  & 26   \\ 
		& & PRODUCT    & 17.93 & 68.42 & 28.42 & 76   & 290  & 52   \\  
		& & LANGUAGE   & 62.96 & 77.27 & 69.39 & 22   & 27   & 17   \\  
		& & WORK\_OF\_ART& 37.32 & 77.11 & 50.29 & 166  & 343  & 128  \\  
		\cmidrule(lr){2-9}
		& \multirow{10}{*}{w/o Chain-of-thought}  
		& PERSON      & 81.44 & 79.23 & 80.32 & 1988 & 1934 & 1575 \\  
		& & ORG        & 69.68 & 69.14 & 69.41 & 1795 & 1781 & 1241 \\  
		& & LOC        & 28.02 & 64.80 & 39.12 & 179  & 414  & 116  \\  
		& & NORP       & 77.73 & 81.33 & 79.49 & 841  & 880  & 684  \\  
		& & GPE        & 96.54 & 89.78 & 93.04 & 2240 & 2083 & 2011 \\  
		& & FAC        & 46.24 & 63.70 & 53.58 & 135  & 186  & 86   \\  
		& & EVENT      & 25.49 & 41.27 & 31.52 & 63   & 102  & 26   \\  
		& & PRODUCT    & 29.71 & 68.42 & 41.43 & 76   & 175  & 52   \\  
		& & LANGUAGE   & 70.37 & 86.36 & 77.55 & 22   & 27   & 19   \\  
		& & WORK\_OF\_ART& 51.00 & 76.51 & 61.20 & 166  & 249  & 127  \\  
		\bottomrule
	\end{tabular}
\end{table*}

\textbf{Named Entity Recognition vs. Named Entity Extraction}

The NER task consists of two sub-tasks: NEE and named entity classification. While previous studies primarily report the overall NER performance, we find that evaluating NEE performance separately can provide deeper insights into model capabilities. As shown in Table~\ref{tb:nee-result}, the strict $F_1$ achieved by all few-shot models in the NEE sub-task are relatively low, ranging from 73.14\% to 88.12\% on CoNLL2003 and from 38.73\% to 74.12\% on OntoNotes5.0. This suggests that the main factor contributing to low strict performance in NER is the low NEE performance, highlighting the need for more focus on improving NEE. Under relaxed match, all few-shot methods perform relatively well on CoNLL2003, with $F_1$ ranging from 90.27\% to 94.32\%. However, they still perform relatively poorly on OntoNotes5.0, with $F_1$ ranging from 60.16\% to 90.63\% in the NEE sub-task. This underscores the need for further improvements in NEE performance to enhance overall NER performance.

Table~\ref{tb:ner-type-detail-result} and Table~\ref{tb:nee-type-detail-result} report detailed metrics—including precision, recall, $F_1$, and counts—for each entity type in both the NER and NEE tasks, comparing GPT4NER with and without the chain-of-thought module. On CoNLL2003, the chain-of-thought helps particularly on complex or ambiguous types such as ORG and MISC, where reasoning over definitions and examples may guide the model to more consistent decisions. Notably, adding the chain-of-thought often increases the number of predicted entities (Pred column) across several types. While this sometimes leads to modest gains in recall, the number of correct predictions (Correct column) does not always increase proportionally. As a result, precision can decrease and $F_1$ may not improve substantially. On OntoNotes5.0, the chain-of-thought similarly increases the number of predicted entities for many types (e.g., PERSON, NORP, WORK\_OF\_ART), but recall gains are limited and precision often drops, indicating that additional reasoning may introduce spurious entities without substantially improving coverage. This suggests that the chain-of-thought can encourage the model to identify more potential entities, but the overall benefit depends on dataset complexity, entity distribution, and context length.

\subsubsection{Ablation Study}\label{sssec:ablation}

\begin{table*}[t]
	\centering
	\caption{Ablation study in the \textbf{NER} task.}
	\label{tb:ner-ablation-result}
	\begin{tabular}{@{}c|c|cccccc@{}}
		\toprule
		\multirow{2}{*}{\textbf{Dataset}}	&	\multirow{2}{*}{\textbf{Method}}	&	\multicolumn{3}{c}{\textbf{Strict Match}}	&	\multicolumn{3}{c}{\textbf{Relaxed Match}}	\\
		\cmidrule(lr){3-5}\cmidrule(lr){6-8}
		&			&	$Pre.$	&	$Rec.$	&	$F_1$	&	$Pre.$	&	$Rec.$	&	$F_1$	\\	
		\midrule
		\multirow{4}{*}{CoNLL2003}	
		&\textbf{GPT4NER}		&\textbf{79.20} &\textbf{87.52}	& \textbf{83.15} & \textbf{81.56} & \textbf{90.12} & \textbf{85.63} \\
		&w/o Entity definition	&	77.64		&	86.99		&	82.05		&	80.50		&	90.19		&	85.07	\\
		&w/o Few-shot examples	&	23.12		&	46.09		&	30.79		&	32.83		&	65.46		&	43.73	\\
		&w/o Chain-of-thought	&	75.57		&	81.82		&	78.57		&	77.66		&	84.08		&	80.74\\
		\midrule
		\multirow{4}{*}{OntoNotes5.0}	
		&\textbf{GPT4NER}		&	67.15	 	&	73.92 	&	70.37		&	79.85		&	87.90		&	83.68 \\
		&w/o Entity definition	&	63.48		&	72.35		&	67.63		&	76.07		&	86.70		&	81.04	\\
		&w/o Few-shot examples	&	36.76		&	45.80		&	40.78		&	44.33		&	55.23		&	49.18	\\
		&w/o Chain-of-thought	&\textbf{71.87}	&\textbf{75.22}	&\textbf{73.50}	&\textbf{84.28}	&\textbf{88.21}	&\textbf{86.20}	\\
		\bottomrule
	\end{tabular}
\end{table*}

\begin{table*}[t]
	\centering
	\caption{Ablation study on the \textbf{NEE} task.}
	\label{tb:nee-ablation-result}
	\begin{tabular}{@{}c|c|cccccc@{}}
		\toprule
		\multirow{2}{*}{\textbf{Dataset}}	&	\multirow{2}{*}{\textbf{Method}}	&	\multicolumn{3}{c}{\textbf{Strict Match}}	&	\multicolumn{3}{c}{\textbf{Relaxed Match}}	\\
		\cmidrule(lr){3-5}\cmidrule(lr){6-8}
		&			&	$Pre.$	&	$Rec.$	&	$F_1$	&	$Pre.$	&	$Rec.$	&	$F_1$	\\	
		\midrule
		\multirow{4}{*}{CoNLL2003}	
		&\textbf{GPT4NER}		& 	83.93 	& 	92.74 	& 	88.12		&	88.13 	& \textbf{97.38}& 	92.52\\
		&w/o Entity definition	&	82.08		&	91.96		&	86.74		&	86.66		&	97.10		&	91.58	\\
		&w/o Few-shot examples	&	31.90		&	63.60		&	42.49		&	46.67		&	93.04		&	62.16	\\
		&w/o Chain-of-thought	&\textbf{86.03}	&\textbf{93.15}	&\textbf{89.45}	&\textbf{89.84}	&	97.27		&\textbf{93.41}	\\
		\midrule
		\multirow{4}{*}{OntoNotes5.0}	
		&\textbf{GPT4NER}		&	70.72		&	77.85		&	74.12		&	86.48		&\textbf{95.20}	& 	90.63 \\
		&w/o Entity definition	&	66.54		&	75.84		&	70.89		&	83.36		&	95.02		&	88.81\\
		&w/o Few-shot examples	&	41.17		&	51.29		&	45.67		&	51.49		&	64.14		&	57.13	\\
		&w/o Chain-of-thought	&\textbf{75.58}	&\textbf{79.11}	&\textbf{77.30}	&\textbf{90.64}	&	94.87		&\textbf{92.71}\\
		\bottomrule
	\end{tabular}
\end{table*}

In our ablation study, we analyze the impact of each component in GPT4NER by systematically removing them one at a time and observing the model performance on both NER and NEE tasks. The best-performing configurations serve as baselines for these experiments. Specifically, GPT4NER with POS tags serves as the baseline for CoNLL2003, while GPT4NER without POS tags serves as the baseline for OntoNotes5.0. The results of these ablation experiments for the NER task are presented in Table~\ref{tb:ner-ablation-result}, and the results for the NEE task are reported in Table~\ref{tb:nee-ablation-result}.

\textbf{Impact of Few-Shot Examples.} Table~\ref{tb:ner-ablation-result} illustrates the significant impact of few-shot examples on the performance of GPT4NER in the NER task. When these examples are removed, the $F_1$ of GPT4NER drop substantially by 52.38 points under strict match and 41.90 points under relaxed match on CoNLL2003, and by 29.59 points under strict match and 34.50 points under relaxed match on OntoNotes5.0. Similarly, Table~\ref{tb:nee-ablation-result} shows that such $F_1$ in NEE decrease by 45.63 points under strict match and 30.36 points under relaxed match on CoNLL2003, and by 28.45 points under strict match and 33.50 points under relaxed match on OntoNotes5.0. These notable decreases in $F_1$ across both matches, tasks, and datasets underscore the critical role of few-shot examples in the performance of GPT4NER.

Conversely, without few-shot examples, GPT4NER essentially operates as a zero-shot model. The results indicate that the introduction of just few-shot examples with minimal human effort can lead to substantial performance improvements in both NER and NEE tasks.

Furthermore, despite the explicit specification of the output format in this experiment, the absence of the implicit output format in the examples led to some generated results having correct content but incorrect format. This inconsistency affects the reliability of subsequent evaluation, as illustrated below:

\begin{itemize}
	\item \textit{Test Text}: This is Xu Li .
	
	\textit{Gold label}: ``Xu Li'': ``\ner{PERSON}''
	
	\textit{Prediction}:
	
	1. \textbf{Xu Li $|$ True $|$ Xu Li is a proper name, making it a PERSON entity.}\\
	2. This $|$ False $|$ This is a pronoun and is excluded from entities.\\
	3. is $|$ False $|$ Is is a verb and is excluded from entities.
	
\end{itemize}

In this example, ``Xu Li'' is correctly identified and classified, but the output does not follow format requirements, and the labeling fails in the subsequent processing.

\textbf{Impact of Entity Definitions.} Table~\ref{tb:ner-ablation-result} reveals that removing entity definition from GPT4NER results in a slight decline in $F_1$ performance for the NER task: a decrease of 1.10 points under strict match and 0.56 points under relaxed match on CoNLL2003, and a decrease of 2.74 points under strict match and 2.64 points under relaxed match on OntoNotes5.0. Similarly, Table~\ref{tb:nee-ablation-result} shows that the $F_1$ scores for NEE decrease by 1.38 points under strict match and 0.94 points under relaxed match on CoNLL2003, and by 3.23 points under strict match and 1.82 points under relaxed match on OntoNotes5.0. These decreases indicate that entity definition is beneficial for both NER and NEE tasks in GPT4NER. However, these $F_1$ decreases are relatively minor compared to the significant drops caused by removing few-shot examples. A possible reason is that LLMs like GPT-3.5 can infer full or partial entity definitions from the input few-shot examples. This suggests that LLMs possess the capability to deduce abstract concepts from specific instances.

\textbf{Impact of Chain-of-Thought.} Table~\ref{tb:ner-ablation-result} shows that removing the chain-of-thought component from GPT4NER leads to a decrease in $F_1$ performance in NER by 4.58 points under strict match and 4.89 points under relaxed match on CoNLL2003. Conversely, without the chain-of-thought component, GPT4NER's performance increases by 3.13 points under strict match and 2.52 points under relaxed match on OntoNotes5.0. Table~\ref{tb:nee-ablation-result} further indicates that without the chain-of-thought component, GPT4NER achieves consistent increases in $F_1$ performance for the NEE task by 1.33 points under strict match and 0.89 points under relaxed match on CoNLL2003, and by 3.18 points under strict match and 2.08 points under relaxed match on OntoNotes5.0. These mixed results suggest that chain-of-thought prompting can be both beneficial and detrimental for few-shot models in NER and NEE tasks. A possible reason for this inconsistency is that chain-of-thought prompting might inadvertently accumulate errors. This implies that while chain-of-thought prompting has potential, its effective design remains challenging and is not always advantageous. The Chain-of-Thought module is originally designed to improve the interpretability of the model, but this module requires the model to add a reason for determining the entity type in the output, which really increases the output complexity of the model.

\begin{table}[t]
	\centering
	\caption{Distribution of POS tags in CoNLL2003 and OntoNotes5.0 datasets, shown as percentage of total entities.}
	\label{tb:pos-tag-distribution}
	\begin{tabular}{@{}c|c|c@{}} 
		\toprule
		\textbf{Dataset} & \textbf{POS tag} & \textbf{Percent.} \\
		\midrule
		\multirow{5}{*}{CoNLL2003} 
		& NN   & 18.69\% \\
		& FW   & 18.04\% \\
		& NNP  & 10.26\% \\
		& UH   & 9.88\%  \\
		& CD   & 7.14\%  \\
		\midrule
		\multirow{5}{*}{OntoNotes5.0} 
		& NN   & 22.10\% \\
		& FW   & 19.56\% \\
		& UH   & 11.20\% \\
		& NNP  & 7.40\%  \\
		& GW   & 6.24\%  \\
		\bottomrule
	\end{tabular}
\end{table}

\textbf{Impact of Part-of-Speech Tags.} Table~\ref{tb:overall-performance} shows that removing POS tags from GPT4NER leads to a decrease in $F_1$ performance in NER by 1.1 points under strict match and 0.8 points under relaxed match on CoNLL2003. However, it results in an increase of 3.6 points under strict match and 4.2 points under relaxed match on OntoNotes5.0. Similarly, Table~\ref{tb:nee-result} reveals that in the NEE task, the performance of GPT4NER without POS tags decreases by 1.6 points under strict match and 0.6 points under relaxed match on CoNLL2003, but increases by 3.1 points under strict match and 3.2 points under relaxed match on OntoNotes5.0.  These results indicate that POS tags are consistently beneficial for CoNLL2003 across both NER and NEE tasks and both match metrics. Conversely,  they are consistently detrimental for OntoNotes5.0 across both tasks and match metrics. A possible explanation for this discrepancy is that in datasets like CoNLL2003, where entity boundaries are clearer and sentence structures more regular, POS tags provide valuable contextual information. By contrast, in datasets like OntoNotes5.0, where entity boundaries are more ambiguous and sentence structures are more diverse and complex, POS tags may introduce noise that negatively affects model performance. Table~\ref{tb:pos-tag-distribution} provides additional insight into the linguistic differences between the datasets. CoNLL2003 is dominated by tags such as NN, NNP, and CD, which offer clear cues for entity recognition. OntoNotes5.0, by contrast, has a higher proportion of NN and FW, with a wider variety of entity types and more complex sentence structures, reducing the effectiveness of POS information and occasionally introducing noise. These observations suggest that POS tags can be beneficial for datasets with clearer entity boundaries and regular sentence structures, such as CoNLL2003, but may not generalize to datasets with more ambiguous boundaries, complex syntax, or fine-grained labels, such as OntoNotes5.0. Careful consideration of dataset characteristics is thus recommended when incorporating POS features in few-shot NER and NEE tasks.

\subsection{Error Analysis}\label{sssec:error-analysis}

There are three main types of errors in the evaluation of GPT4NER: 

(1) \textbf{Post-Processing Errors}. The addition of POS tags often leads to the omission of spaces around hyphens and possessive markers on OntoNotes5.0, making it difficult to locate the corresponding phrases in test text for annotation, as illustrated below:

\begin{itemize}
	
	\item \textit{Test Text}: Does the President still believe that \textbf{Kim Jong - Il} is a tyrant a pygmy and a spoiled child .
	
	\textit{Gold label}: ``\textbf{Kim Jong - Il}'': ``\ner{PERSON}''
	
	\textit{Prediction}: \textbf{Kim Jong-Il} $|$ True $|$ as it is a person's name (PERSON) 
	
	\item \textit{Test Text}: in Minneapolis Lucy Dalglish executive director of \textbf{the Reporters ' Committee for Freedom of the Press} .
	
	\textit{Gold label}: ``Minneapolis'': ``\ner{GPE}'', ``Lucy Dalglish'': ``\ner{PERSON}'', ``\textbf{the Reporters ' Committee for Freedom of the Press}'': ``\ner{ORG}''
	
	\textit{Prediction}: \textbf{Reporters' Committee for Freedom of the Press} $|$ True $|$ as it is the name of an organization (ORG)
	
\end{itemize}

When processing test texts with POS tags, LLMs tend to focus on the POS of connectors. When generating output, they often omit spaces around connectors to follow formal expressions. However, this can introduce issues for subsequent processing.

(2) \textbf{Hallucination Errors}. GPT4NER occasionally returns entity types that are not included in the entity-definition component. This issue is particularly evident in ablation experiments when few-shot examples are removed, which increases the likelihood of hallucinations in LLMs, as illustrated below:

\begin{itemize}
	
	\item \textit{Test Text}: At present , we should not have a problem with watching television .
	
	\textit{Gold label}: None
	
	\textit{Prediction}: ``present'': ``\ner{TIME}'', ``problem'': ``\ner{PROBLEM}'', ``television'': ``\ner{PRODUCT}''
	
	\item \textit{Test Text}: And let me go back to January of two thousand two in the President 's axis of evil speech before congress .
	
	\textit{Gold label}: ``congress'': ``\ner{ORG}''
	
	\textit{Prediction}: ``January'': ``\ner{DATE}'', ``two thousand two'': ``\ner{DATE}'', ``President'': ``\ner{TITLE}'', ``axis of evil'': ``\ner{PHRASE}'', ``congress'': ``\ner{ORG}''
	
\end{itemize}

Hallucinations often occur in experiments where POS tags are added or few-shot examples are removed. POS tags introduce additional complexity, and without few-shot examples, the model's learning becomes less robust.

(3) \textbf{Annotation Errors}. These errors often stem from annotator oversight. Despite rigorous review and proofreading, minor mistakes are inevitable, as illustrated below:

\begin{itemize}
	
	\item \textit{Test Text}: JAPAN GET LUCKY WIN, \textbf{CHINA} IN SURPRISE DEFEAT
	
	\textit{Gold label}: ``JAPAN'': ``\ner{LOC}'', ``\textbf{CHINA}'': ``\ner{PER}''
	
	\item \textit{Test Text}: Rumsfeld: The Iraqis received us with overwhelming happiness and welcomed us, because of the practices of your bloody regime over the course of all those years in which you governed \textbf{Iraq}.
	
	\textit{Gold label}: ``Rumsfeld'': ``\ner{PERSON}'', ``Iraqis'': ``\ner{NORP}'', ``\textbf{Iraq}'': ``\ner{PERSON}''
	
\end{itemize}

Obviously, ``CHINA'' and ``Iraq'' refer to countries or locations instead of persons.

\subsection{Limitations}\label{ssec:limitations}

There are two primary limitations in our work. The first concerns interpretability. Depending solely on ChatGPT's reasoning within candidate entities may not provide sufficient interpretability. LLMs such as ChatGPT introduce uncertainty in their generated output, and explanations provided in the results may not always be accurate or reliable. The second limitation pertains to token constraints. While using larger models like GPT-4 and GPT-5 may enhance performance, this is not our main focus.

\section{Conclusion}\label{sec:conclusion}

This paper introduces GPT4NER, a method based on LLMs for few-shot NER. GPT4NER constructs effective prompts using entity definition, few-shot examples, chain-of-thought, and POS tags to leverage the capabilities of LLMs in transforming the few-shot NER task into a sequence-generation task. Experimental results on two benchmark datasets demonstrate that GPT4NER significantly outperforms state-of-the-art few-shot models and achieves competitive results compared to fully-supervised models. Furthermore, our experiments advocate for the use of the relaxed-match metric (which is widely used in time expression recognition and normalization) to evaluate model performance. Additionally, our experiments also suggest to report performance in the NEE sub-task to deepen insights into model capabilities in the NER task.

\bibliography{aaai2026}

\end{document}